\documentclass[letterpaper]{article} % DO NOT CHANGE THIS
\usepackage{aaai2027}   % DO NOT CHANGE THIS
\usepackage[hyphens]{url}  % DO NOT CHANGE THIS
\usepackage{graphicx} % DO NOT CHANGE THIS
\usepackage{natbib}  % DO NOT CHANGE THIS AND DO NOT ADD ANY OPTIONS TO IT
\usepackage{caption} % DO NOT CHANGE THIS AND DO NOT ADD ANY OPTIONS TO IT
\usepackage{algorithm}
\usepackage{algorithmic}

\usepackage[utf8]{inputenc} % allow utf-8 input
\usepackage[T1]{fontenc}    % use 8-bit T1 fonts
\usepackage{url}            % simple URL typesetting
\usepackage{booktabs}       % professional-quality tables
\usepackage{amsfonts}       % blackboard math symbols
\usepackage{nicefrac}       % compact symbols for 1/2, etc.
\usepackage{microtype}      % microtypography
\usepackage{xcolor}         % colors
\usepackage{amsmath}

\usepackage{tcolorbox}
\usepackage{adjustbox}
\usepackage{graphicx}
\usepackage{subcaption} 
\usepackage{booktabs}
\usepackage{siunitx}
\usepackage{xcolor}
\usepackage{caption}
\usepackage{subcaption}
\usepackage{tabularx}

\usepackage{newfloat}
\usepackage{listings}
\DeclareCaptionStyle{ruled}{labelfont=normalfont,labelsep=colon,strut=off} % DO NOT CHANGE THIS
\DeclareFloatingEnvironment[name=Listing,placement=tb,fileext=lst]{listing}
\lstdefinestyle{json}{%
    basicstyle={\small\ttfamily},numbers=none,xleftmargin=0pt,%
    breaklines=true,breakatwhitespace=false,breakautoindent=true,%
    columns=fullflexible,keepspaces=true,showstringspaces=false,%
    stringstyle=\color[rgb]{0.73,0.13,0.13},string=[b]",%
    literate={\{}{{\{}}1 {\}}{{\}}}1 {[}{{[}}1 {]}{{]}}1}

\usepackage{booktabs}

\title{LITMUS — \underline{L}iterature-\underline{I}nformed \underline{T}estbed for \underline{M}apping \underline{U}se-cases and \underline{S}ynthesizing Risk-Cards}
\author{
    Written by AAAI Press Staff\textsuperscript{\rm 1}\thanks{With help from the AAAI Publications Committee.}\\
    AAAI Style Contributions by Peter Patel Schneider,
    Sunil Issar,\\
    J. Scott Penberthy,
    George Ferguson,
    Hans Guesgen,
    Francisco Cruz\equalcontrib\corresponding,
    Marc Pujol-Gonzalez\equalcontrib\corresponding
}
\affiliations{
    \textsuperscript{\rm 1}Association for the Advancement of Artificial Intelligence\\
    1101 Pennsylvania Ave, NW Suite 300\\
    Washington, DC 20004 USA\\
    proceedings-questions@aaai.org
}

\author {
   Seshu Tirupathi\textsuperscript{\rm 1}
   Elizabeth M. Daly\textsuperscript{\rm 1}, 
   Radu Marinescu \textsuperscript{\rm 1},
   Javier Carnerero Cano\textsuperscript{\rm 1},
   Yannis Belkhiter \textsuperscript{\rm 1}
}
\affiliations {
    \textsuperscript{\rm 1}IBM Research\\
    seshutir@ie.ibm.com, elizabeth.daly@ie.ibm.com, radu.marinescu@ie.ibm.com, javier.cano@ibm.com, yannis.belkhiter@ibm.com
}
\begin{document}

\maketitle

\begin{abstract}
The proliferation of large language models (LLMs) and Artificial Intelligence (AI) in real‑world applications has heightened the need for systematic, safety‑oriented deployment frameworks. Existing approaches typically prioritize the construction of datasets and risk cards around the LLM and AI architecture itself, yet they provide scant consideration for the specific use‑case that drives model selection. This omission is problematic because any practical AI deployment begins with a detailed use‑case description that informs the assessment of contextual hazards. In this paper, we introduce a structured pipeline that, given a use‑case, automatically produces a tailored dataset and an accompanying risk card. The process ensures that identified risks are tightly coupled to the use‑case, and it offers traceable evidence from peer‑reviewed literature to substantiate each risk to use‑case relationship. Prior claim-verification methods (FActScore, FactReasoner) have been validated for factual claims. Risk-relevance is a hypothetical claim about possible harm, and whether evidence retrieval can meaningfully assess such claims remains untested. Literature-Informed Testbed for Mapping Use-cases and Synthesizing (LITMUS) risk cards are the first systematic attempt to characterize this adaptation. We evaluated the pipeline using the MIT Risk Taxonomy and achieved a 0.91 F1-score with a list of 320 samples from the benchmark risk dataset that was considered. These results demonstrate the effectiveness of our approach in generating robust, evidence‑based risk assessments for LLM deployments.
\end{abstract}
% \begin{links}
%     \link{Code}{https://anonymous.4open.science/r/litmus-7503/}
% \end{links}

\section{Introduction}
\label{sec:intro}
Large language models (LLMs) are increasingly employed across a broad spectrum of applications, from conversational agents \cite{liao2023proactive} and virtual assistants \cite{guan2023intelligent} to automated code synthesis \cite{zhou2026automating} and beyond. Although the automation capabilities of AI yield evident benefits, the frequency of AI incidents reported in connection with their deployment has increased proportionally \cite{feffer2023ai}. These incidents span a wide risk continuum, ranging from minor, low‑consequence events that minimally affect stakeholders to severe, potentially catastrophic failures. In light of the expanding portfolio of LLM and AI use‑cases and the corresponding escalation of associated risks, it is essential to devise a structured methodology that evaluates the risks inherent to a particular LLM or AI application prior to deployment. Such a methodology would facilitate the identification of appropriate guardrails and mitigation strategies, thereby ensuring that LLMs are integrated in a safe and responsible manner.

Historically, the governance, risk assessment, and management of LLMs has been conducted manually, a process that is both time‑consuming and susceptible to errors \cite{celestin2025future}. Recent initiatives have sought to accelerate this workflow by automating portions of the governance cycle, for example by zero-shot LLMs, RAG based LLMs \cite{salwala2024user} or employing few‑shot or Chain‑of‑Thought (CoT) LLMs to parse compliance questionnaires. Nonetheless, these approaches remain highly sensitive to the selection of few‑shot examples and the design of CoT prompts. Moreover, while few‑shot strategies can produce plausible risk evaluations, they typically lack transparent source attribution, hindering rigorous scrutiny of the identified risks. The lack of systematic, evidence‑anchored risk inventories and absence of a reusable artifact for generating risks for a given use-case provides the motivation for this work. 

 As part of this effort we draw on recent work in automated claim verification (\cite{thorne2018fever}, \cite{min2023factscore}, and FactReasoner \cite{marinescu2025factreasoner}), which decomposes generated text into atomic propositions and grounds each in retrieved evidence. These methods were developed and validated for factual claims: statements that are  true, false, or unsupported relative to documented fact. Risk-relevance claims are not of this type. The statement "unfair discrimination is a risk for an AI-powered vehicle speech-command system" is a hypothetical claim about possible harm: a judgment about what could go wrong, for whom, under what conditions, grounded partially in documented incidents and partially in reasoned  extrapolation from system properties. A risk that has never manifested and has no literature trace may nonetheless be the most important to flag, a limitation we return to in Appendix \ref{app:limitations}.  Whether evidence-retrieval methods designed for factual verification can meaningfully assess such hypothetical claims is, to our knowledge, untested. LITMUS is structured around this question: we adapt atomic-claim verification for risk-relevance claims, and characterise where the transfer succeeds (high concordance with documented incidents in the AI Incident Database) and where it strains (risk categories with sparse literature, and the asymmetry between absence-of-evidence and evidence-of-absence).

Claim-verification methods, including those that explicitly reason over conflicting or uncertain sources such as FactReasoner, are designed to assess propositions with empirical truth conditions: claims about what is, was, or has been documented to occur. The available evidence may be incomplete or contradictory, but the underlying claim is in principle empirically resolvable. Hypothetical risk claims are not of this form. The statement that "risk Y is relevant for use-case X" has no equivalent empirical truth condition. Two equally informed experts can legitimately disagree on whether a particular risk applies to a particular use-case, because the judgment requires both evidence and value-laden choices about which harms to consider. We do not claim LITMUS resolves this by producing ground-truth risk labels. We claim however we can provide structured evidence retrieval which can make the empirical component of risk judgments transparent, traceable, and reproducible, leaving the judgment component for human review. This is a smaller claim than "automated risk assessment" but a substantively stronger one than "expert-curated risk catalog," because it gives downstream practitioners (auditors, regulators, deployment teams) a defensible audit trail from each (use-case, risk) association to the documented evidence that supports or contradicts it.

In this context, we make the following contributions through this work:
\noindent
\begin{itemize}
     \item A modular, open-source framework with a pluggable evidence backend.
    The default configuration retrieves contexts at query time from public
    web sources via Google Search, but the same pipeline can be re-instantiated
    against any curated corpus (e.g., a vector-indexed collection of
    peer-reviewed scientific literature or policy documents), enabling
    reproducible evaluation against credibility-controlled evidence bases.
    \item A methodology that adapts atomic-claim verification
    (\textsc{FactReasoner}) to risk-relevance assessment.
    % \item A benchmark of \(720\) pairs of anonymised AI use-case annotated with
    % risk associations from the MIT AI Risk Repository taxonomy and supporting
    % evidence atoms and a further 269 use-cases validated for concordance against the AI Incident Database.
    \item The use-case risk card artifact for documenting evidence-grounded
    (use-case, risk) associations with traceable source attribution. 
\end{itemize}

\section{Literature Review}
\label{sec:related-work}
LITMUS sits at the intersection of three research areas. Risk governance and taxonomy researchers \cite{ai2024artificial}, \cite{smuha2025regulation}, \cite{slattery2026ai}, \cite{zeng2024ai},\cite{bagehorn2025ai}  have produced systematic enumerations of AI risk types, but the mapping from abstract risk type to specific use-case context is left to undocumented expert judgment. AI safety benchmark researchers  \cite{mazeika2024harmbench, ghosh2025ailuminate, wang2023decodingtrust, han2024wildguard} have produced large-scale evaluations of model behaviour under adversarial prompts, but they evaluate what a model produces rather than which risks are relevant to a particular application. Claim-verification researchers (\cite{thorne2018fever}, \cite{min2023factscore}, and FactReasoner \cite{marinescu2025factreasoner}) have produced methods for grounding statements in retrieved evidence, but for factual claims rather than for the hypothetical risk-relevance claims central to risk assessment. None of these threads directly addresses the question of which risks are relevant to which AI applications, backed by what evidence, which is the gap LITMUS aims to address.

Detailed literature review is presented in the Appendix. 

% Text-mining for safety claims is an active field applying NLP to extract, classify, and analyze risk factors from unstructured text, such as incident reports, maintenance logs, and social media \cite{ZHANG2019,akhbardeh2020,agarwal2025}.
\section{Methodology}
\label{sec:methodology}
\begin{figure*}[h]
    \centering
    \includegraphics[width=0.5\linewidth]{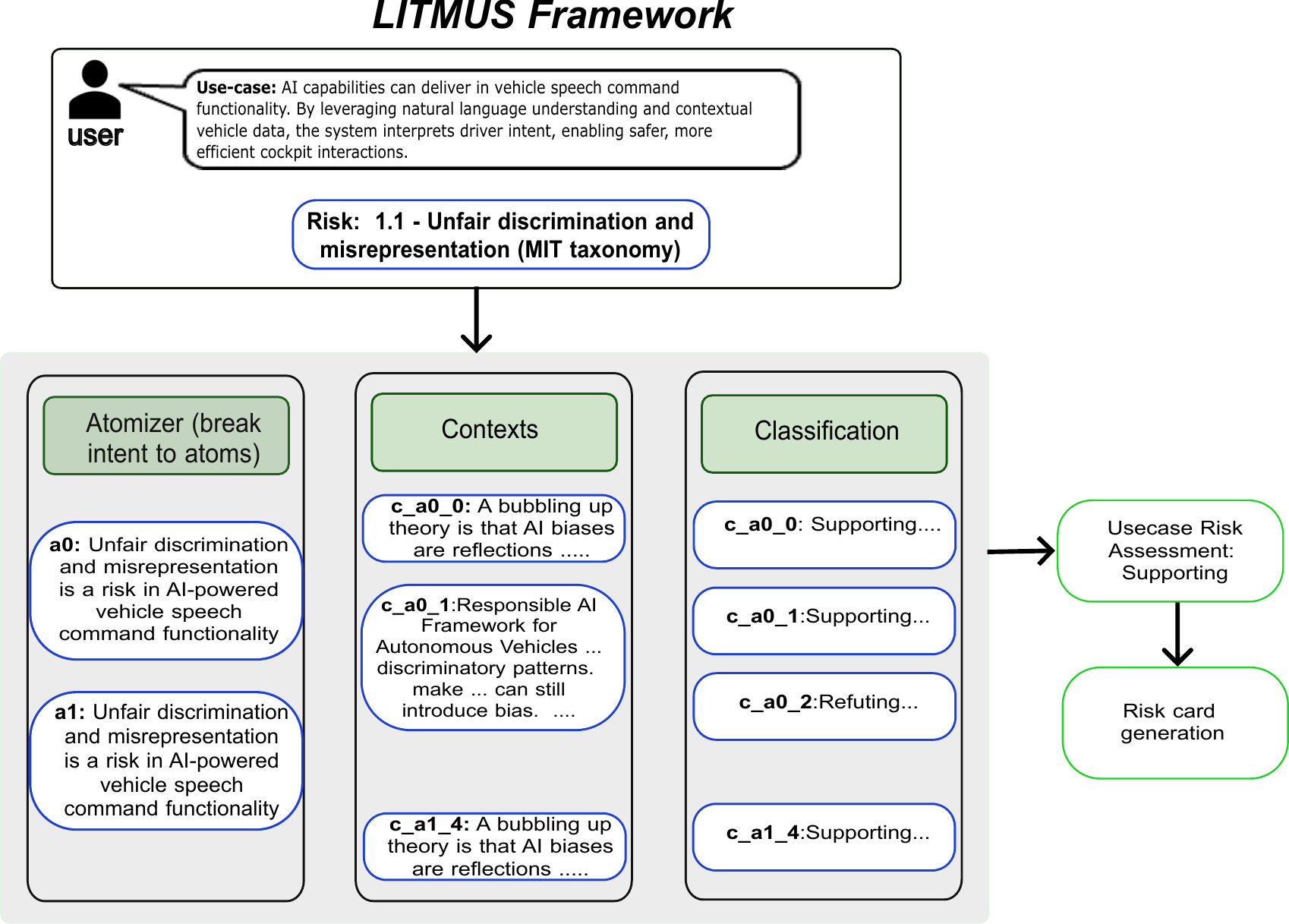}
    \caption{LITMUS architecture diagram of risk card generation from scientific literature for AI use-cases.}
    \label{fig:arch}
\end{figure*}

LITMUS frames risk-relevance assessment as an instance of long-form factuality evaluation adapted to the (use-case, risk) setting. We build on the \textsc{FactReasoner} architecture~\cite{marinescu2025factreasoner}, a neuro-symbolic factuality assessor that decomposes a long-form response into atomic units, retrieves contextual evidence, and uses a probabilistic graphical model to estimate the posterior probability that each atom is supported. The methodological extension that distinguishes LITMUS is a \emph{Coupling} stage that transforms a hypothetical risk claim into a factually-shaped input that the \textsc{FactReasoner} pipeline can assess unchanged. The full pipeline is illustrated in Fig.~\ref{fig:arch}.

Let $u \in \mathcal{U}$ denote a natural-language use-case description and $r \in \mathcal{T}$ a risk type drawn from a taxonomy $\mathcal{T}$ (we use the MIT AI Risk Repository Domain Taxonomy~\cite{slattery2026ai}). We define the \emph{risk-relevance claim} $\varphi(u, r)$ as the proposition that risk $r$ is relevant for use-case $u$, expressed in natural language as
``\emph{$r$ is a risk in $u$}''. Given an external knowledge source $\mathcal{K}$, the LITMUS objective is to compute a verdict $v(u, r) \in \{\textsc{supporting}, \textsc{contradicting}\}$ grounded in evidence retrieved from $\mathcal{K}$. As discussed in the Introduction section, $\varphi(u, r)$ is a hypothetical claim about possible harm rather than a factual proposition. LITMUS characterises only the empirical component of risk relevance, namely whether retrieved evidence supports the claim, and leaves the judgmental component for human review.

\textsc{FactReasoner} \cite{marinescu2025factreasoner} is a recent factuality assessor that, unlike previous prompt-based approaches \cite{min2023factscore,song2024veriscore,wei2024long}, leverages probabilistic reasoning to assess the factuality of the LLM generated response with respect to an external knowledge source $\mathcal{K}$. Specifically, the \textsc{FactReasoner} pipeline consists of four stages called \emph{Atomizer}, \emph{Reviser}, \emph{Retriever} and \emph{Evaluator}, respectively. The \emph{Atomizer} decomposes the input response $y$ into a set of $n$ atomic units $\mathcal{A}_y$ by applying any of the decomposition strategies proposed recently \cite{min2023factscore,bayat2025factbench}. The \emph{Reviser} post-processes the atoms such that the pronouns, unknown entities, or incomplete names are replaced with their corresponding named entities in the response \cite{wei2024long}. Next, the \emph{Retriever} is responsible for querying the external knowledge source $\mathcal{K}$ to retrieve the contexts relevant to the response's atoms \cite{song2024veriscore}. Finally, the \emph{Evaluator} constructs a graphical model representing a joint probability distribution over the atomic units in the response and their corresponding contexts in $\mathcal{K}$. For each atom $a_i$, it then computes the posterior marginal probability $P(a_i)$ which quantifies the likelihood that $a_i$ is true (or supported) given the information available in $\mathcal{K}$. Finally, the factuality metrics such as factual precision $Pr(y)$, recall  and $F1@K$, can be readily calculated using the supported atoms \cite{marinescu2025factreasoner}. Note that both the \emph{Atomizer} and \emph{Reviser} stages use LLMs for their specific tasks, as previously shown in \cite{marinescu2025factreasoner}.

\textsc{FactReasoner} takes a single long-form response $y$ as input and assesses its factual content against $\mathcal{K}$. LITMUS instead takes a $(u, r)$ pair and constructs the risk-relevance claim
\begin{equation}
    \varphi(u, r) = \text{``\emph{r} is a risk in \emph{u}''},
    \label{eq:coupling}
\end{equation}
which is then passed to the \textsc{FactReasoner} \emph{Atomizer} in place of $y$. For example, with $u = $ ``\emph{AI capabilities can deliver in-vehicle speech-command functionality\ldots}'' and $r = $ MIT 1.1 (Unfair discrimination and misrepresentation), the coupled claim is ``\emph{Unfair discrimination and misrepresentation is a risk in AI-powered vehicle speech-command systems}''. The Atomizer then decomposes this claim into atomic sub-claims of the form $a_i = $ ``\emph{r} is a risk in \emph{$u_i$}'', where $u_i$ is a sub-component or sub-task identified within $u$. Each atom isolates a specific (sub-use-case, risk) pair so that retrieved evidence can be evaluated against a self-contained empirical sub-claim.

Once the risk-relevance claim $\varphi(u, r)$ is coupled, LITMUS applies the \textsc{FactReasoner} pipeline directly. The use-case level verdict is obtained by thresholding the factual precision score
$Pr(\varphi(u, r))$:
\begin{equation}
v(u, r) = \begin{cases}
    \textsc{supporting} & \text{if } Pr(\varphi(u, r)) > \tau \\
    \textsc{contradicting} & \text{otherwise}
\end{cases}
\label{eq:verdict}
\end{equation}

with $\tau = 0.5$ throughout. The verdict and the underlying $Pr(\varphi(u, r))$ score together populate the use-case risk card
(discussed in Use-case Risk Card Section).

\textsc{FactReasoner} distinguishes three atom-level outcomes \cite{marinescu2025factreasoner}: \emph{supported} ($P(A_i) > P(\neg A_i)$), \emph{contradicted} ($P(\neg A_i) > P(A_i)$), and \emph{undecided} ($P(A_i) \approx P(\neg A_i) \approx 0.5$, indicating insufficient information to resolve the atom). At the use-case level, LITMUS treats only the \emph{supported} outcome as evidence in favour of $\varphi(u, r)$, and collapses both \emph{contradicted} and \emph{undecided} into a single \textsc{contradicting} verdict. This collapses \emph{evidence-of-absence} and \emph{absence-of-evidence} into a single non-supporting verdict.

\begin{table*}[htbp]
    \centering
    \caption{Definitions of classification metrics for both positive and negative claims.}
    \label{tab:class_metrics}
    \begin{tabular}{ll}
        \toprule
        \textbf{Metric} & \textbf{Definition} \\
        \midrule
      True Positive (TP)  &  Risk truly exists and model predicts it\\
      False Negative (FN) & Risk truly exists and model predicts none \\
      False Positive (FP) & Risk does not exist and model predicts it \\
    True Negative (TN) & Risk does not exist and model predicts none \\
        \bottomrule
    \end{tabular}
\end{table*}

\section{Data Description: Use-cases and Risk Taxonomies}
\label{sec:dataset}
% For the present study, we considered the set of use‑cases supplied by Google \footnote{https://cloud.google.com/transform/101-real-world-generative-ai-use-cases-from-industry-leaders} 
% as a benchmark for evaluating risk‑identification performance. Before any analysis, these use‑cases were anonymized by removing proprietary details. The anonymization process involved prompting a large language model (LLM) with a handful of few‑shot examples and explicit instructions to produce anonymized versions. These use-cases will be referred to as Benchmark Dataset for the remaining paper. The few‑shot examples and instruction template are displayed in Fig.~\ref{fig:prompt_use_cases} (Appendix). A subset of the intents are provided in Fig.~\ref{fig:example_use_cases} (Appendix). 

% Additionally, 
To evaluate the fact‑reasoning component against examples with known ground truth, we employed incidents sourced from the AI Incident Database (AIID) \cite{mcgregor2021preventing}. These incidents were anonymized, and a large language model (LLM) was instructed using a set of few‑shot examples to transform each incident into a generic use‑case. The transformation assumes that the event has not yet occurred, thereby eliminating bias from the original incident description and allowing a clean assessment of whether a specific risk is present or absent. The few‑shot examples and the instruction template used for this conversion are illustrated in Fig. \ref{fig:prompt_incident} (Appendix) and a subset of the generated intents are presented in Fig.~\ref{fig:example_incidents} (Appendix).

Further, Since there are no incident databases where a given risk is explicitly stated as not relevant to a given use-case, we used ChatGPT to generate a synthetic dataset covering all 24 MIT risk categories from the MIT Risk Taxonomy. For each risk, we produced five unrelated scenarios, yielding 120 “no‑risk” pairs (e.g., Unfair discrimination and misrepresentation is not a risk with a lane‑change maneuver system that only activates when sensor thresholds are met).

Evidence for risk-relevance atoms is retrieved at query time from publicly accessible web sources via Google Search using the Serper API, following the retrieval design of \textsc{FactReasoner}~\cite{marinescu2025factreasoner}. Retrieved sources include peer-reviewed scientific articles, policy and standards documents, technical reports, and industry analyses. We do not pre filter the corpus by source type or credibility. The framework also supports an offline mode in which retrieved documents are persisted to a ChromaDB vector index, enabling reproducible runs against a fixed evidence base of the user's choosing; we discuss this option and its implications for credibility control in Section A (Appendix).

Finally, we choose the MIT risk taxonomy for the experiments (Table \ref{tab:mit_risk_taxonomy}). 
\section{Experiments}
\label{sec:experiments}
In this section, we describe the experiments to benchmark and corroborate our approach. We analyze data drawn from the AI Incident Database and synthetic dataset, comparing the supporting and non‑supporting arguments produced by LITMUS framework against the recorded incidents and compare them with GPT-OSS. 
% Second, we test the framework's performance on the Benchmark Dataset and assess the accuracy of atom generation, supporting risk generation, and entailment prediction by benchmarking them against ground‑truth labels curated through human feedback.

\subsection{Corroboration with AI Incident Database and Synthetic Dataset}

We evaluated two prompt templates as described below on each of the 120 synthetic pairs and 200 AIID samples:
\begin{itemize}
    \item \textbf{Positive claim}: "[RISK] is a risk associated with [USECASE]"
Ex. Unfair discrimination and misrepresentation is a risk associated with a lane‑change maneuver system that is triggered only when sensor arrays report distances and speeds that satisfy explicit safety thresholds.
\item \textbf{Negative claim}: "[RISK] is not a risk associated with [USECASE]"
Ex. Unfair discrimination and misrepresentation is not a risk associated with a lane‑change maneuver system that is triggered only when sensor arrays report distances and speeds that satisfy explicit safety thresholds.
\end{itemize}

Here, [RISK] and [USECASE] are substituted from the the MIT Risk Taxonomy and AIID incidents and synthetic dataset samples (Table \ref{tab:mit_risk_taxonomy} and Table \ref{tab:synthetic_risks} in the Appendix). The positive and negative claims check that the model and the risk relevance generated are not sensitive to the statement provided in the prompt itself. 

We compare the performance of GPT-OSS-120B and the LITMUS framework where consistent definitions (Table \ref{tab:class_metrics}) are used as evaluation metrics for both models and prompts. 
% We employ 269 anonymized use‑case descriptions extracted from the AIID corpus (as described in Section \ref{sec:dataset}), each accompanied by the risk category assigned to it in AIID. These descriptions and their associated risk labels are then processed through the LITMUS pipeline, using the risk label and the anonymized use‑case text as the input. Across the 269 use‑cases extracted from AIID, only 0.6\% of the generated arguments are classified as non‑supporting (Fig.~\ref{fig:aiid_distribuion_atoms}). Moreover, only a single incident is judged non‑supporting as can be seen in (Fig.~\ref{fig:aiid_distribuion_incidents}); all remaining incidents are correctly matched to their respective risk categories, and the supporting literature provided by LITMUS framework corroborates the existence of each identified risk.
The classification performance is summarised in Table \ref{tab:class_report_positive} and Table \ref{tab:class_report_negative}. GPT‑OSS‑120B and LITMUS show similar performance for positive prompts while GPT‑OSS‑120B shows a markedly higher false‑positive rate when asserting that a risk does not apply in the prompt, whereas LITMUS remains far more consistent. This shows that LITMUS is not sensitive to the claim and looks for supporting and non-supporting arguments for a given claim where as GPT‑OSS‑120B has a bias towards predicting that a risk exists for any given use-case. Further, GPT‑OSS‑120B often cites hallucinated sources (Table~\ref{tab:gpt_citations}) while LITMUS pulls citations directly from Google Scholar via the SERPER API, eliminating such hallucinations.
Finally, GPT‑OSS‑120B supplies only supporting arguments to support the prediction that was made. In contrast, LITMUS presents both supporting and opposing evidence where available, giving users a clearer context for each risk–use‑case pairing (Example in Listing~\ref{listing:entailment_contradiction}).

These results confirm that LITMUS framework reliably identifies supporting or contradicting arguments, accurately classifies the presence or absence of a given risk category, and provides verifiable evidence for the use-case-risk pairs in the AIID and the synthetic dataset samples.
\begin{table*}[htbp]
    \centering
    \small
    \caption{Classification metrics for risk identification for positive claim.}
    \label{tab:class_report_positive}
    \begin{tabular}{l cccc cccc}
        \toprule
        \textbf{Model} & \textbf{TP} & \textbf{FP} & \textbf{FN} & \textbf{TN} & \textbf{Accuracy} & \textbf{Precision} & \textbf{Recall} & \textbf{F1-Score} \\
        \midrule
        GPT-OSS-120B & 199 & 32 & 1 & 88 & 0.89 & 0.86 & 0.99 & 0.92 \\
        LITMUS       & 196 & 33 & 4 & 87 & 0.88 & 0.86 & 0.98 & 0.91 \\
        \bottomrule
    \end{tabular}
\end{table*}

\begin{table*}[htbp]
    \centering
    \small
    \caption{Classification metrics for risk identification for negative claim.}
    \label{tab:class_report_negative}
    \begin{tabular}{l cccc cccc}
        \toprule
        \textbf{Model} & \textbf{TP} & \textbf{FP} & \textbf{FN} & \textbf{TN} & \textbf{Accuracy} & \textbf{Precision} & \textbf{Recall} & \textbf{F1-Score} \\
        \midrule
        GPT-OSS-120B & 200 & 55 & 0 & 65 & 0.82 & 0.78 & 1.00 & 0.87 \\
        LITMUS       & 177 & 11 & 23 & 109 & 0.94 & 0.94 & 0.89 & 0.91 \\
        \bottomrule
    \end{tabular}
\end{table*}

\begin{table*}[htbp]
    \centering
    \small
    \caption{Citation accuracy - 15 samples (75 inferences of GPT-OSS-120B). Human evaluation: for each entries, we evaluate the correctness of the supporting citation provided by the model. Namely, we check: the name of the report/paper, the authorship, the year of the publication, the venue and the URL (or DOI if scientific paper).}
    \label{tab:gpt_citations}
    \begin{tabular}{lll}
        \toprule
        \textbf{Metric} & \textbf{Value} & \textbf{Definition} \\
        \midrule
Rows checked	&  75	&15 use-cases × 5 sample citations each \\ 
Paper name correct (exact match)	&33.3\%	& Only 1 in 3 citations matches an actual paper title \\
All 4 reference fields correct (author and year and venue and DOI/URL)	&18.7\% &	Fewer than 1 in 5 citations is fully clean end-to-end \\
Avg. fraction of fields correct (per-entry mean)	&33.0\% &	On average, only ~1.3 of the 4 fields are right per entry \\
— Author accuracy	&38.7\% &	\\
— Year accuracy	& 38.7\% &	\\
— Venue accuracy	& 30.7\% &	\\
— DOI/URL accuracy	& 24.0\%	& Weakest field — most fabricated component\\
        \bottomrule
    \end{tabular}
\end{table*}

To flesh out the incorrect conclusion of LITMUS framework for AIID, Listing \ref{listing:contradiciton_example_aiid} illustrates the case that yields a contradictory assessment. The anonymized use‑case concerns a model that forecasts housing prices (Listing \ref{listing:contradiciton_example_aiid} in Appendix), whereas AIID has the actual incident as “Zillow shutting down its Offers‑Division due to inaccurate price predictions”\footnote{https://incidentdatabase.ai/cite/149/} and the corresponding risk identified as insufficient capability or robustness. When the LITMUS framework processes the anonymized description and the corresponding risk, the literature‑search component surfaces scholarly sources that substantiate the existence of highly accurate house‑price prediction models. Consequently, the evidence retrieved by LITMUS directly contradicts the risk reported by AIID.

\subsection{Human Verification}

Finally, we conducted a study to gauge the alignment between the LITMUS judgments and expert assessments for three risk‑analysis questions posed on a representative sample of 50 JSON records. Each record contains an intent statement, a risk scenario (atom), an explanatory risk‑text, and a label indicating whether the atom is entailed by the risk-text (Fig.~\ref{fig:questionnaire} Appendix). Three domain experts independently answered the following three questions for each record, selecting from the drop‑down options whether the option was correct, incorrect or unclear. We considered three questions:
\begin{enumerate}
    \item Is the risk scenario (atom) logically consistent with the stated intent?
    \item Is the supporting risk text relevant to this specific risk scenario (atom)?
    \item Is the relation type ("entailment"/"contradiction") correct?
\end{enumerate}

The annotation protocol involved a random sample of records from the benchmark datasets with risks listed in the MIT Risk taxonomy. Inter‑annotator agreement was quantified using pairwise percent agreement and then averaging the score which is $72\%$ across all annotator pairs (Table \ref{tab:annotator_agreement_base}) which indicates a reasonable agreement overall. The relevance and entailment\_check metrics have a moderate agreement which indicates that the annotators were more divergent in their judgment of these two categories. 
% The resulting majority vote (2 of 3) served as the ground‑truth reference against which the algorithm’s predictions were compared.

  \begin{table}[!ht]                                                              
  \centering          
  \small
  \caption{Pairwise Percent Agreement Between Annotators}
  \label{tab:annotator_agreement_base}
  \begin{tabular}{|l|c|c|c|c|}                                                  
  \hline                                                                        
  \textbf{Factor} & \textbf{1-2} & \textbf{1-3} & \textbf{2-3} &
  \textbf{Mean} \\                                                              
  \hline          
  Consistency & 84.00\% & 90.00\% & 86.00\% & 86.67\% \\                            
  \hline                                                                        
  Relevance & 60.00\% & 74.00\% & 68.00\% & 67.33\% \\
  \hline                                                                        
  Entailment Check & 62.00\% & 64.00\% & 60.00\% & 62.00\% \\
  \hline                                                                        
  \textbf{Average} & 68.67\% & 76.00\% & 71.33\% & 72.00\% \\
  \hline                                                                        
  \end{tabular}   
  \end{table}  

Table \ref{tab:algorithm_performance} reports the per‑question and overall accuracy of the LITMUS algorithm considering the ground truth based on majority voting of the three annotators. For logical consistency (Q1) the algorithm achieved 98\% accuracy (49/50 matches), for risk‑text relevance (Q2) 80.9\%, and for entailment type correctness (Q3) 85.7\% accuracy. The overall agreement across all questions was 88.4\% (129/146 matches). These figures indicate a strong correspondence with majority expert judgment. 

  \begin{table*}[h]                                                              
  \centering
  \caption{LITMUS Performance vs Human Consensus (Majority Vote)}            
  \label{tab:algorithm_performance}                                             
  \begin{tabular}{lcccccc}
  \toprule                                                                      
  \textbf{Feedback Type} & \textbf{Total} & \textbf{Consensus} & \textbf{3-0} & 
  \textbf{2-1} & \textbf{Matches} & \textbf{Accuracy} \\                        
  \midrule                                                                      
  Consistency & 50 & 50 & 40 & 10 & 49/50 & 98.0\% \\                           
  Relevance & 50 & 47 & 27 & 20 & 38/47 & 80.9\% \\
  Entailment & 50 & 49 & 22 & 27 & 42/49 & 85.7\% \\                            
  \midrule                                                                      
  \textbf{Overall} & \textbf{150} & \textbf{146} & \textbf{89} & \textbf{57} &  
  \textbf{129/146} & \textbf{88.4\%} \\                                         
  \bottomrule     
  \end{tabular}                                                                 
  \end{table*}

In summary, the human‑feedback evaluation demonstrates that the algorithm performs at a level comparable to expert reviewers, with high percentage of agreement with the majority votes. Further tests on inter-annnotator agreement are presented in Section H (Appendix). 

\begin{listing}
  \begin{lstlisting}[style=json,basicstyle={\footnotesize\ttfamily}]
{
    "Use-case Description": "<Use-case description>",
    "Risk Taxonomy": "<Choice of Risk Taxonomy, ex: MIT>",
    "Relevant Risks": "<Risks from the risk taxonomy relevant to the use-case description>",
    "Documentation": "<LITMUS generated scientific literature that provides evidence for the existence of the risks mentioned in relevant risks>",
    "Additional Information": "<Additional information on the risks and supporting documentation. ex. knowledge graph of scientific literature or risk probabilities.>"
}
\end{lstlisting}
  \caption{Use-case risk card template which provides information on the risks associated with a use-case along with supporting documentation.}
  \label{listing:usecase_risk_card}
\end{listing}

\begin{listing}
\begin{lstlisting}[style=json,basicstyle={\footnotesize\ttfamily}]
{
    "Risk Definition": "<Sub-category from MIT Risk Taxonomy>",
    "Risk Taxonomy": "<Choice of Risk Taxonomy, ex: MIT>",
    "Use-case Description": "Use-case description",
    "Documentation": "<LITMUS generated scientific literature that provides evidence for the existence or non-existence of the risk mentioned in risk definition>",
    "Additional Information": "<Additional information on the risk and documentation. ex. knowledge graph of scientific literature or risk probabilities.>"
}
\end{lstlisting}
  \caption{Risk card template which provides information on the supporting or non-supporting documentation associated with a use-case.}
  \label{listing:risk_usecase_card_template}
\end{listing}

\section{Use-case Risk Card}
\label{sec:risk-card}
Building on the documentation tradition reviewed in the related work, we propose a use-case-level extension. The closest existing artifact in this space is the Impact Assessment Card \cite{bogucka2025impact}, which targets the \emph{communication} of AI risks and benefits to diverse stakeholder audiences through a visual format. LITMUS risk cards address a complementary problem: \emph{populating} the data underlying such artifacts with traceable, evidence-grounded (use-case, risk) associations. The two could be deployed in a complementary manner, with LITMUS providing the evidence base and the Impact Assessment Card providing the audience-facing presentation. Understanding the applicability of a system to a given use-case is crucial, because the use-case description is the first step in designing an AI solution, yet a standardised use-case risk card would streamline the capture of all relevant threats and would benefit from algorithms that automatically populate its key-value pairs.  In this paper, we present two automated approaches for generating risk-use‑case cards (Listings \ref{listing:usecase_risk_card}, \ref{listing:risk_usecase_card_template}), thereby normalizing the way risks are identified and documented for any given use-case:
\begin{itemize}
    \item Risk enumeration – For a given use-case, list the relevant risks drawn from a predefined taxonomy, together with supporting literature, beginning with the use‑case description.
    \item Evidence retrieval – For each risk–use‑case pair, fetch supporting or non‑supporting documentation from LITMUS.
\end{itemize}
These methods enable consistent, reproducible risk documentation across diverse AI deployments. Concrete example risk cards are shown in Listings \ref{listing:usecase_risk_card_example}, \ref{listing:risk_usecase_card_example} in Appendix \ref{app:risk_cards}.

\section{Conclusion}
We presented a systematic, literature‑driven framework that generates risk cards for specific LLM use‑cases. We first decompose each use‑case into atomic intents that pairs every atomic AI intent with a risk from the risk taxonomy. Using these pairs, we search the scientific literature, retrieve contexts that address both the risk and the use‑case, and classify each context as either supporting or non‑supporting the risk’s presence.

We applied this method to a collection of anonymised AI use‑cases. We compared our results to the AI Incident Database and verified a random subset with human experts, achieving high concordance. More fundamentally, LITMUS provides the first systematic test of whether evidence-retrieval methods designed for factual claims can be meaningfully adapted to hypothetical claims about possible harm. We find that this transfer succeeds for well-documented risk categories and strains for sparsely-documented ones, suggesting that the empirical component of risk-relevance judgments can be made transparent, traceable, and reproducible at the use-case level for the documented portion of the risk landscape. By leveraging our pipeline, we provide researchers and practitioners with a concrete, evidence‑based path to understand and document the risks that any LLM deployment might entail for a given use-case. Future work will focus on improving the inter-annotator agreement for risk text relevance and entailment checks. 

% %%%%%%%%%%%%%%%%%%%%%%%%%%%%%%%%%%%%%%%%%%%%%%%%%%%%%%%%%%%%

\bibliography{aaai2027}

% Check whether the conference requires a reproducibility checklist to be included in the paper.
% If so, you can uncomment the following line and ajust the path to include it.
% \input{ReproducibilityChecklist.tex}

\newpage
\appendix
\onecolumn

\appendix
\section{Appendix}
\subsection{A. Extensions}
\label{app:extensions}

\begin{figure}[htbp]
    \centering
    \begin{subfigure}{0.45\textwidth}
        \includegraphics[width=\linewidth]{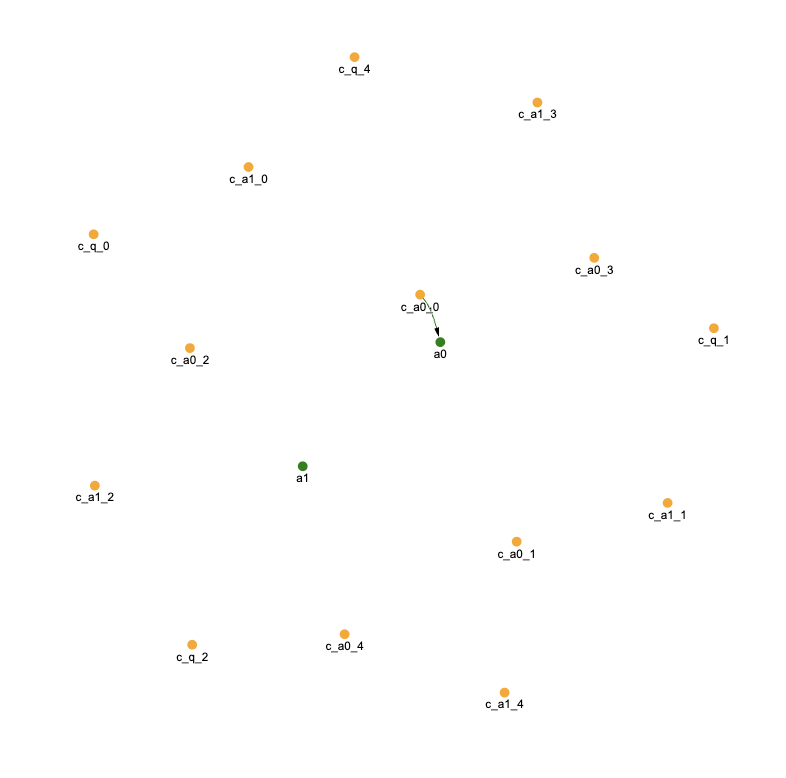}
        \caption{Competitive dynamics risk}
    \end{subfigure}
    \hfill % Adds horizontal space
    \begin{subfigure}{0.45\textwidth}
        \includegraphics[width=\linewidth]{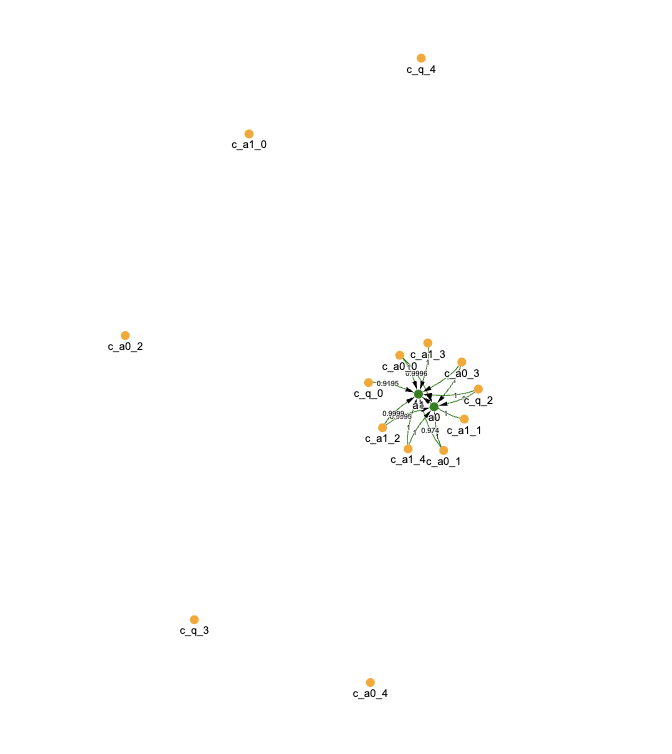}
        \caption{Unfair discrimination and representation}
    \end{subfigure}
    \caption{Comparison of risk support for two subcategories in MIT Risk Taxonomy for the use-case: "AI capabilities can deliver in vehicle speech command functionality. By leveraging natural language understanding and contextual vehicle data, the system interprets driver intent, enabling safer, more efficient cockpit interactions."}
    \label{fig:risk_support}

\end{figure}

The present study reports the primary outcomes obtainable with the algorithm illustrated in Fig.~\ref{fig:arch}. This framework serves as a versatile foundation that can be adapted to a range of scenarios in which the description of a use‑case within the AI‑management lifecycle varies. The following four extensions demonstrate how the LITMUS pipeline can be tailored to accommodate different end‑user constraints:
\begin{itemize}
    \item \textbf{Refined risk‑relevance estimation}: Our initial method assigns a set of risks to a use‑case, but a more sophisticated algorithm could yield a finer‑grained assessment of risk relevance (e.g., high, medium, low). For instance, the current approach identifies both “competitive dynamics” and “unfair discrimination/representation” as relevant for the use‑case “AI‑enabled in‑vehicle speech‑command functionality.” Although correct, Fig.~\ref{fig:risk_support} reveals that the supporting contexts for competitive dynamics are substantially fewer than those for discrimination/representation. Incorporating metrics that capture atom density and the connectivity of their supporting contexts would enhance the precision of risk relevance attribution.
    \item \textbf{Pluggable evidence backend and corpus substitution.}
    The default LITMUS configuration uses live Google Search as the evidence
    backend, which provides broad coverage of scientific, policy, and industry
    sources but offers no credibility filtering and no reproducibility guarantee
    across re-runs. Because the \textsc{Retriever} stage of the pipeline is
    modular, the live backend can be replaced by any retriever that returns
    contexts in the standard $(t, l, s, d)$ tuple form (title, link, snippet,
    document content). Two substitutions are particularly relevant for
    high-stakes risk assessment use-cases:
\begin{enumerate}
    \item \textit{Curated peer-reviewed corpora.} Persisting retrieved
    documents in a ChromaDB vector index allows users to restrict the
    evidence base to peer-reviewed scientific articles or policy documents
    that meet their credibility criteria. Recent systematic literature
    reviews provide ready-made corpora of this kind: for instance,
    \cite{lee2026constructing} curate a set of 112 primary studies on AI
    safety cases assembled through systematic database search and citation
    snowballing, each scored against quality-assessment criteria. Such a
    corpus could be used directly as $\mathcal{K}$, producing risk
    assessments grounded only in vetted academic sources at the cost of
    coverage breadth.
    \item \textit{Domain-specific knowledge bases.} Sectoral evidence bases
    (e.g., medical-device incident reports, automotive safety standards,
    financial-services regulatory filings) can be used when the use-case
    falls within a regulated domain and domain-specific evidence is more
    appropriate than the generalist web. This is particularly relevant for
    deployments subject to sector-specific compliance regimes.
\end{enumerate}
The trade-off across backends is between coverage (live web search surfaces
emerging risks earlier and includes grey literature) and credibility
(curated corpora restrict to vetted sources but lag the current literature
and exclude policy and industry analyses). We do not advocate one backend
over another in general; the appropriate choice depends on the downstream
use of the risk card and the credibility threshold required by the consuming
audience (auditor, regulator, deployment team).

    \item \textbf{Compatibility with alternative risk taxonomies}: The atom‑generation stage requires only a high‑level risk definition, making it straightforward to substitute or extend the current taxonomy choice. The MIT Risk Taxonomy used in our experiments is one of the frameworks integrated by Risk Atlas Nexus \cite{bagehorn2025ai}, and LITMUS could be re-instantiated against any of the taxonomies that framework consolidates (\cite{mit,owasp,zeng2024ai}).
    \item \textbf{Dynamic risk‑card adaptation to use‑case changes}: Leveraging use‑case similarity metrics and knowledge‑graph representations, the LITMUS framework can assess whether previously identified risks remain applicable when a use‑case description is modified. This capability enables rapid re‑evaluation and re‑generation of risk cards based on similarity measures rather than a full re‑analysis.
\end{itemize}
Together, these extensions illustrate the flexibility of the LITMUS pipeline and its potential to support a broad spectrum of risk‑assessment needs across diverse AI deployments.

\subsection{B. Limitations}
\label{app:limitations}

\begin{figure}[h]
    \centering
    \includegraphics[width=0.9\columnwidth]{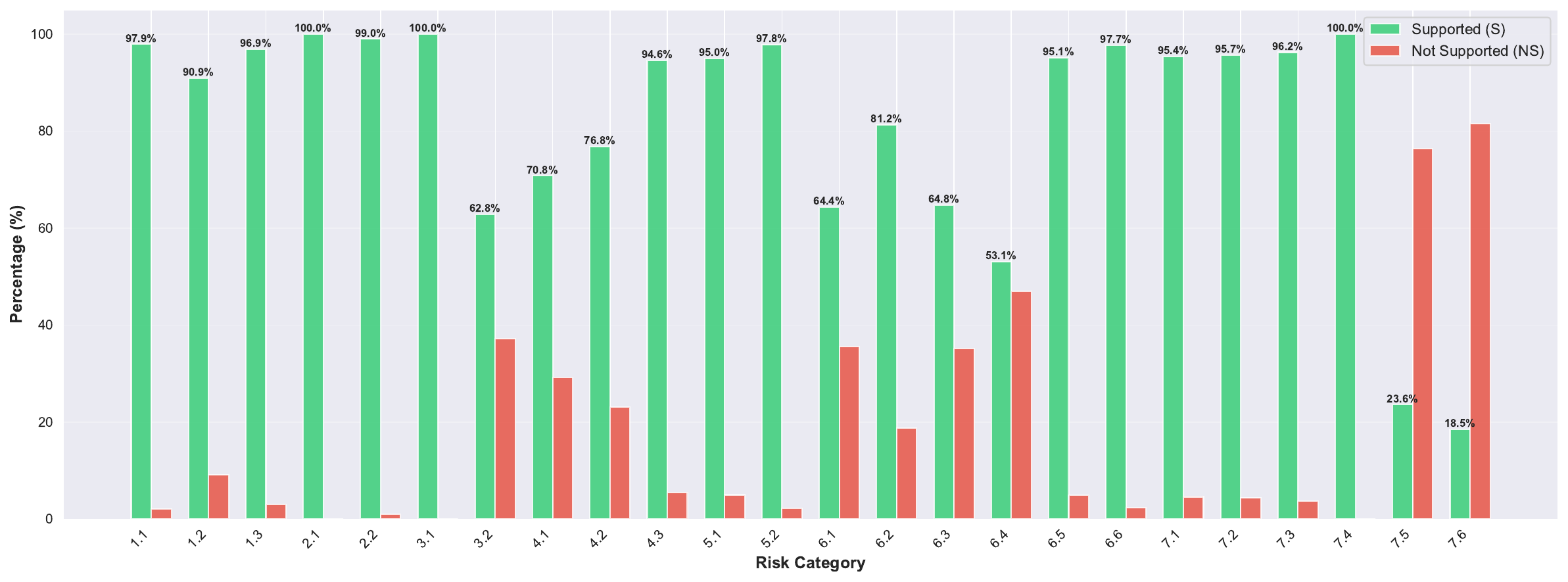}
    \caption{Distribution of supported and non-supported contexts by risk subcategory in a benchmark dataset of 269 LLM use-cases. }
    \label{fig:context_support}
\end{figure}

Although our approach performs well, we identified some constraints that users should consider before deployment or further improvement:
\begin{itemize}
    \item \textbf{Novel and emerging risks}. LITMUS surfaces risks for which retrievable evidence exists. Genuinely novel hazards, those that have not yet manifested in deployed systems, or that have manifested  but are not yet documented in peer-reviewed literature, incident databases, or governance reports, will receive few or no supporting atoms and will be classified as non-supporting. This is a structural property of evidence retrieval, which is by construction oriented toward what has been documented rather than toward what could occur. The "Multi agent risks (7.6)" subcategory of the MIT Risk Taxonomy illustrates this Fig.~\ref{fig:context_support}, where the available literature is substantially sparser than for established categories such as "Unfair discrimination," producing more non-supporting classifications even when the risk is genuinely applicable to the use-case. We therefore do not claim LITMUS provides a complete enumeration of risks for a given use-case. Its strongest use is to provide reproducible, traceable grounding for the documented portion of the risk landscape, alongside expert anticipation of emerging hazards rather than as a substitute for it.   
    \item \textbf{Scope of the risk cards} The method reports only whether a risk is relevant to a given use‑case; it does not quantify the risk’s expected impact or likelihood. Moreover, it fails to differentiate how the relevance varies across stakeholder groups—such as end users, developers, or platform providers. 
    \item \textbf{Handling of marginally relevant contexts} When all retrieved contexts receive a relevance probability of 0.5, the algorithm treats the atom as “non‑supporting.” Yet this outcome can also signify a genuine absence of literature that explicitly confirms or refutes the risk for that use‑case. We recommend interpreting such cases as evidence gaps rather than definitive exclusions.
    \item \textbf{Dependence on use‑case descriptions} The algorithm relies on the richness of the use‑case narrative. If the description remains terse, we capture fewer supporting contexts from the literature, thereby limiting our risk insight. We therefore advise that use‑case descriptions explicitly outline the AI tasks involved to maximize risk extraction.
    \item \textbf{Credibility of retrieved literature} Our current pipeline does not rate source credibility; consequently, we might ingest contexts from low‑trust publications. We can mitigate this limitation by applying a post‑processing step that scores each source before classifying contexts as supporting or non‑supporting. Additionally, leveraging an offline, well‑curated database of scientific literature would further guarantee the correctness of the retrieved contexts.
\end{itemize}

% \FloatBarrier 

\subsection{C. Literature Review}
\paragraph{Risk, Regulation, AI Safety} Risk governance for AI systems has developed along two largely parallel tracks: regulatory frameworks inherited from broader risk management traditions and technical AI safety research. From a regulatory perspective, ISO 31000 \cite{iso31000} frames risk as uncertainty over objectives and prescribes iterative risk management processes. AI safety research frames risk as emerging from model–environment interaction, emphasizing robustness and specification errors \cite{amodei2016concrete} and motivating alignment methods such as learning from human feedback \cite{christiano2017preferences,ouyang2022training}.

\paragraph{Risk Taxonomies} Common AI risk taxonomies organise threats into domains such as bias and fairness, privacy and security, malicious misuse, and technical reliability (e.g., hallucinations and distributional drift) \cite{slattery2026ai, nist-risk}. %Representative frameworks, including the MIT AI Risk Repository \cite{slattery2026ai} and the NIST AI Risk Management Framework \cite{nist-risk}, use these categories to link technical vulnerabilities—such as prompt injection or model collapse—to downstream business impacts, including reputational damage, legal liability, and financial loss.
The MIT AI Risk Repository \cite{slattery2026ai} used in our experiments provides a comprehensive risk catalog, consolidating existing taxonomy sources through expert authorship. %LITMUS uses the MIT Domain Taxonomy in our experiments, and while the MIT AI Risk Repository catalogs which risks have been described in the literature, LITMUS assesses which of those risks are evidentially supported for a particular AI deployment.
Beyond individual taxonomies, recent work has focused on the interoperability problem created by the proliferation of frameworks. Risk Atlas Nexus \cite{bagehorn2025ai, vejsbjerg2026risk} consolidates risks across multiple sources \cite{nist-risk, slattery2026ai, owasp} using ontologies and knowledge graphs, and exposes programmatic methods for risk identification given a use-case description. %This reframes the AI risk taxonomy landscape as an interoperability problem rather than a question of which framework is canonical. LITMUS builds directly on this line of work and extends it: where Risk Atlas Nexus provides taxonomy infrastructure, LITMUS adds the evidence dimension, grounding each (use-case, risk) association in retrievable literature.

\paragraph{AI Safety Benchmarks.} A parallel line of work evaluates AI safety by testing model behaviour under adversarial or risk-eliciting prompts. DecodingTrust \cite{wang2023decodingtrust} introduces multi-dimensional trust evaluation across toxicity, stereotype bias, adversarial robustness, privacy, machine ethics, and fairness. HarmBench \cite{mazeika2024harmbench} provides a standardised framework for automated red-teaming across diverse harm categories. AILuminate  \cite{ghosh2025ailuminate} establishes industry-grade benchmarks for production safety testing.  WildGuard \cite{han2024wildguard} and Granite Guardian \cite{padhi2025granite} operate as inline classifiers that flag unsafe inputs or outputs at deployment time. These benchmarks evaluate the model outputs in response to specific inputs measuring whether a model produces unsafe content when prompted. This is complimentary to the contribution of  LITMUS which addresses which risks are relevant for a particular use-case, supported by what documented evidence. Behavioural benchmarks evaluate whether a model is safe to deploy, LITMUS evaluates which risks deserve documentation and  oversight in a specific deployment. % (HarmBench \cite{mazeika2024harmbench}, AILuminate \cite{ghosh2025ailuminate}, DecodingTrust \cite{wang2023decodingtrust}, WildGuard \cite{han2024wildguard}

\paragraph{Text Mining for Safety Claims} 
Text mining for safety claims has evolved into an NLP subfield that goes beyond sentiment analysis to detect risks, harms, and failures in unstructured text across domains such as AI‑generated content and industrial records. Recent work emphasizes targeted safety signal detection, e.g., identifying "smoke terms" that indicate hazardous conditions in reviews or reports have shown to outperform coarse sentiment cues for surfacing safety issues \cite{goldberg2023}. In industrial settings, NLP models including SVMs, Random Forests, BERT, and hybrids with Bayesian networks are used to extract failure modes and latent risk factors from highly technical maintenance and accident reports \cite{ZHANG2019}. Other methods address bias and toxicity in language through fine‑grained taxonomies \cite{kou2025,gehman2020,nadeem2020,nangia2020}, benchmark datasets \cite{xu2026}, and model‑based evaluators \cite{koh2024} that better capture subtle harms than traditional classifiers. For large language models, these efforts increasingly converge on continuous monitoring and red‑teaming to identify evolving safety risks, reflecting the limitations of static metrics under rapid model iteration \cite{ganguli2022,ge2023}.

\paragraph{Documentation Standards.} A parallel line of work has produced standards for documenting AI artifacts, their evaluations, and their risks. Datasheets for Datasets \cite{gebru2021datasheets},  Model Cards \cite{mitchell2019model}, and Data Cards \cite{pushkarna2022data} introduced structured templates for documenting datasets and trained models. BenchmarkCards \cite{sokol2024benchmarkcards} adapted the format to LLM benchmarks. Closer to the present work, AI Risk Profiles \cite{sherman2024ai} propose a standardized template for pre-deployment risk disclosure, and the Impact Assessment Card \cite{bogucka2025impact} introduces a visual artifact for communicating AI risks and benefits to diverse stakeholder audiences, empirically validated through user studies showing improved task completion and output quality compared to traditional impact reports. A complementary structural-argument tradition adapts safety case methodology from classical engineering to AI systems \cite{lee2026constructing}, providing templates for evidence-grounded safety arguments. These artifacts aim to improve transparency through structured documentation for downstream accountability. LITMUS extends this paradigm to use-case-level risk assessment, producing risk cards that document which risks attach to a deployment and the retrieved evidence supporting each pairing.

\subsection{D. Datasets Generation}

\begin{figure}[!ht]
\centering
\footnotesize
\begin{adjustbox}{max width=\textwidth}
\begin{tcolorbox}[colback=gray!5, colframe=black, title=Prompt Template, fonttitle=\bfseries]

Instructions:
Your task is to produce a given incident into a rephrased version which does not have any company information or proprietary information and should focus only on the AI tasks of the original use-case prior to this incident occuring.

Rules:
- The rephrased version should be 2-3 sentences long.
- The rephrased version must describe only the AI tasks in the paragraph.
- The rephrased version should ignore the actual incident that occured and just focus on the AI tasks. 
- The rephrased version must not have any company or proprietary information.
- The rephrased version must be generic and not include numbers which can also be proprietary. 
- The output must be a JSON dictionary with one key and the rephrased version:

json
{
    "id1": "<rephrased version>",
}

Use the provided examples to learn your task.

Example 1:

INPUT: * YouTube’s content filtering and recommendation algorithms exposed children to disturbing and inappropriate videos.,Google’s YouTube Kids App Presents Inappropriate Content

OUTPUT: json
{
    "id1": "A system designed to analyze video content is intended to recommend new videos to users, providing a curated experience. It aims to filter the content of suggested videos, ensuring only suitable options are presented. The system also seeks to help users discover new and relevant content based on their interests."
}

Example 2:

INPUT: * A self-driving public shuttle by Keolis North America and Navya was involved in a collision with a human-driven delivery truck in Las Vegas, Nevada on its first day of service.

OUTPUT: json
{
    "id1": "A self-driving public shuttle is designed to provide safe and efficient transportation to passengers by harnessing advanced autonomous technology. It aims to reduce congestion and emissions while improving mobility for the elderly and disabled. The vehicle operates independently, making trips between designated stops with precision and reliability."
}

Example 3:

INPUT: * Study on database reports of robotic surgery malfunctions (8,061), including those ending in injury (1,391) and death (144), between 2000 and 2013.

OUTPUT: json
{
    "id1": "Researchers plan to design a database analysis tool to monitor robotic surgery complications, with the goal of identifying and preventing serious issues. The study's primary objective is to track and report on malfunctions, injuries, and fatalities related to robotic-assisted surgical procedures. This data-driven approach aims to improve the safety and efficacy of these procedures."
}

\end{tcolorbox}
\end{adjustbox}
\caption{Prompt used to generate the intents for the incidents in AI Incident Database. }
\label{fig:prompt_incident}
\end{figure}

\begin{figure}[!ht]
\centering
\footnotesize
\begin{adjustbox}{max width=\textwidth}
\begin{tcolorbox}[colback=gray!5, colframe=black, title=Prompt Template, fonttitle=\bfseries]

INPUT:  Microsoft's Tay, an artificially intelligent chatbot, was released on March 23, 2016 and removed within 24 hours due to multiple racist, sexist, and anti-semitic tweets generated by the bot.,
.

OUTPUT: 
{
    "id1": "A chatbot system is designed to engage in conversation with users, generating human-like responses to a wide range of topics and questions. It aims to learn from interactions and adapt its responses to better serve users, providing helpful and informative answers. The system seeks to simulate natural language understanding and generation, enabling effective communication with users."
}

INPUT: Wikipedia bots meant to remove vandalism clash with each other and form feedback loops of repetitve undoing of the other bot's edits.

OUTPUT: json
{
    "id1": "Automated editing tools are designed to maintain the integrity of online content by detecting and reverting malicious changes. These tools aim to preserve the accuracy and reliability of information by correcting errors and removing inappropriate edits. The systems work to identify and revert unwanted changes, ensuring a stable and trustworthy environment for users."
}

INPUT: Uber vehicles equipped with technology allowing for autonomous driving running red lights in San Francisco street testing.

OUTPUT: json
{
    "id1": "A system utilizing autonomous technology is designed to enable vehicles to navigate through streets safely and efficiently. It aims to detect and respond to traffic signals, ensuring compliance with traffic rules and regulations. The technology seeks to improve road safety by reducing the risk of accidents caused by human error."
}

\end{tcolorbox}
\end{adjustbox}
\caption{Example AI incidents and rephrased incident use-cases.}
\label{fig:example_incidents}
\end{figure}

\subsection{E. Benchmark Dataset}
In this section, we consider the benchmark dataset with 269 LLM use-cases (provided in Github) and test the applicability of the LITMUS framework to understand if a given risk is associated with a given use-case.
This analysis examines context (literature) repetition patterns across 24 MIT risk taxonomy subcategories, analyzing 720 JSON documents containing 8,159 unique context sources. 

\begin{figure}[h]
    \centering
    \includegraphics[width=0.9\columnwidth]{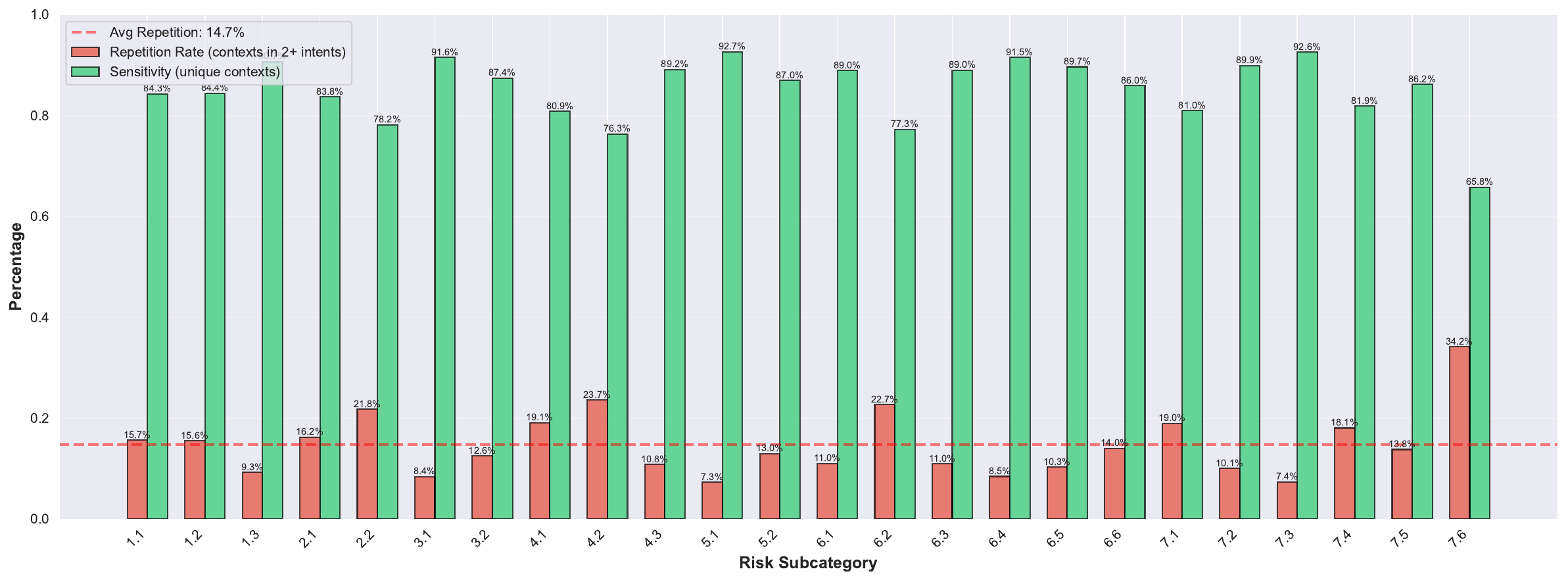}
    \caption{Context Repetition vs Sensitivity by Risk Subcategory.}
    \label{fig:repetition_mit}
\end{figure}

\begin{figure*}[htbp]
    \centering
    \begin{subfigure}{0.32\textwidth}
        \includegraphics[width=\linewidth]{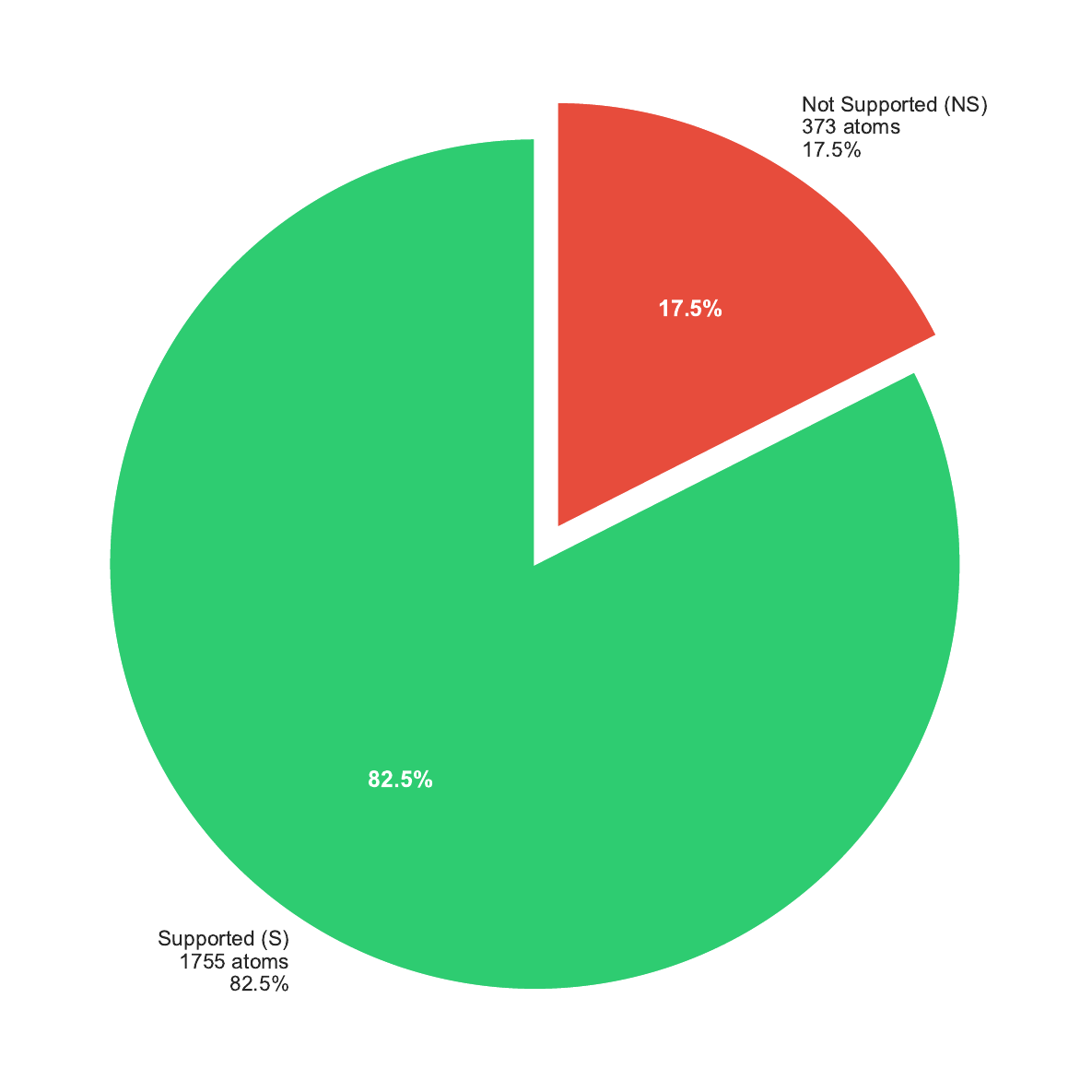}
        \caption{Distribution of supported and not supported atoms  for Benchmark Dataset.}
        \label{fig:mit_distribuion_atoms}
    \end{subfigure}
    \hfill % Adds horizontal space
    \begin{subfigure}{0.52\textwidth}
        \includegraphics[width=\linewidth]{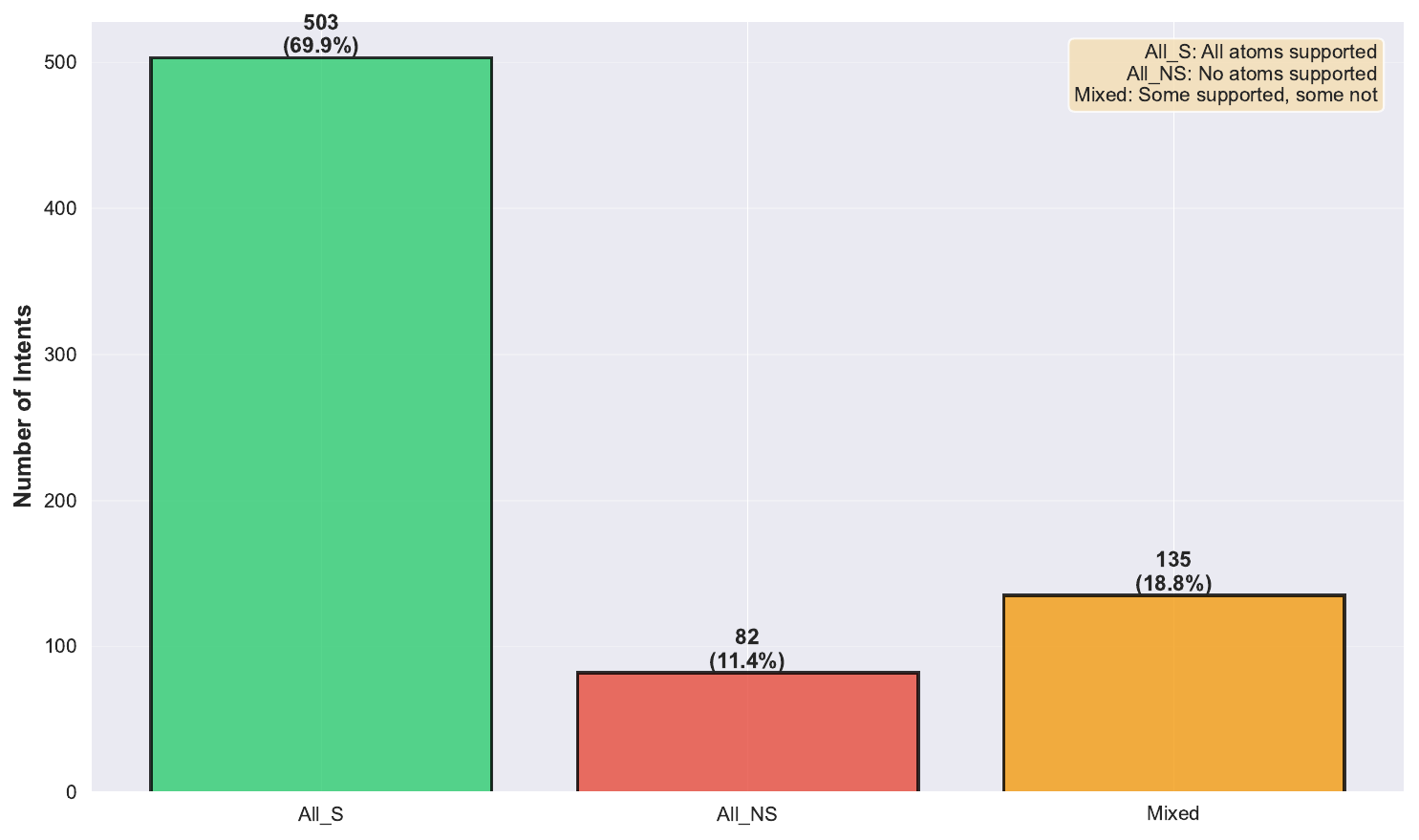}
        \caption{Distribution of supported and not supported incidents for Benchmark Dataset.}
        \label{fig:mit_distribuion_intents}
    \end{subfigure}
    \caption{LITMUS distribution metrics for Benchmark Dataset.}
    \label{fig:litmus_intent}
\end{figure*}

On average, only 14.7\% of contexts repeat within a subcategory, meaning 85.3\% of contexts are unique to individual use-case scenarios, indicating strong retrieval diversity and sensitivity to the atoms extracted from the use-case (Fig.~\ref{fig:repetition_mit}). Further, Fig.~\ref{fig:mit_distribuion_atoms} illustrates the number of supporting and non-supporting contexts for the various MIT risk taxonomy sub categories. The sizeable number of non-supported atoms and categories in Fig.~\ref{fig:mit_distribuion_intents} can also be attributed to the limited literature available for emergent risk categories like Multi-agent risks (7.6 in Fig.~\ref{fig:context_support}). This information can also be used to expand on the work for these subcategories in the future work. 

\section{Risk Cards}
\label{app:risk_cards}
\begin{listing}[!ht]
\begin{lstlisting}[style=json]
{
    "Use-case Description": "AI capabilities can analyze vast amounts of data to provide real-time insights for optimizing fleet operations and improving driver safety. By processing large datasets, the system enables data-driven decision making for more sustainable and safer transportation systems.",
    "Risk Taxonomy": "MIT",
    "Relevant Risks": "1.2",
    "Supporting Documentation": "[{1.2: Toxic content is likely present in AI training data scraped from public sources, which could pose a risk in AI-based applications, including potentially AI-based transportation.}]",
    "Additional Information": "[probability: 0.98]"
}
\end{lstlisting}
  \caption{Example use-case risk card which provides information on the risks associated with a use-case along with supporting documentation.}
  \label{listing:usecase_risk_card_example}
\end{listing}

\begin{listing}[!ht]
\begin{lstlisting}[style=json]
{
    "Risk Definition": "1.2 - Exposure to toxic content", 
    "Risk Taxonomy": "MIT",
    "Use-case Description": "AI capabilities can analyze vast amounts of data to provide real-time insights for optimizing fleet operations and improving driver safety. By processing large datasets, the system enables data-driven decision making for more sustainable and safer transportation systems.",
    "Documentation": "[Scientific literature actual text: 3. Toxicity. AI training data scraped from public websites, user forums, and social media pages is highly likely to contain toxic content.]",
    "summary": "Toxic content is likely present in AI training data scraped from public sources, which could pose a risk in AI-based applications, including potentially AI-based transportation.",
    "Additional Information": "[probability: 0.98]"
}
\end{lstlisting}
  \caption{Example risk card which provides information on the supporting or non-supporting documentation associated with a use-case.}
  \label{listing:risk_usecase_card_example}
\end{listing}

\subsection{F. Human Verification}
\label{app:human_verification}

\begin{figure}[!ht]
    \centering
    \includegraphics[width=0.7\columnwidth]{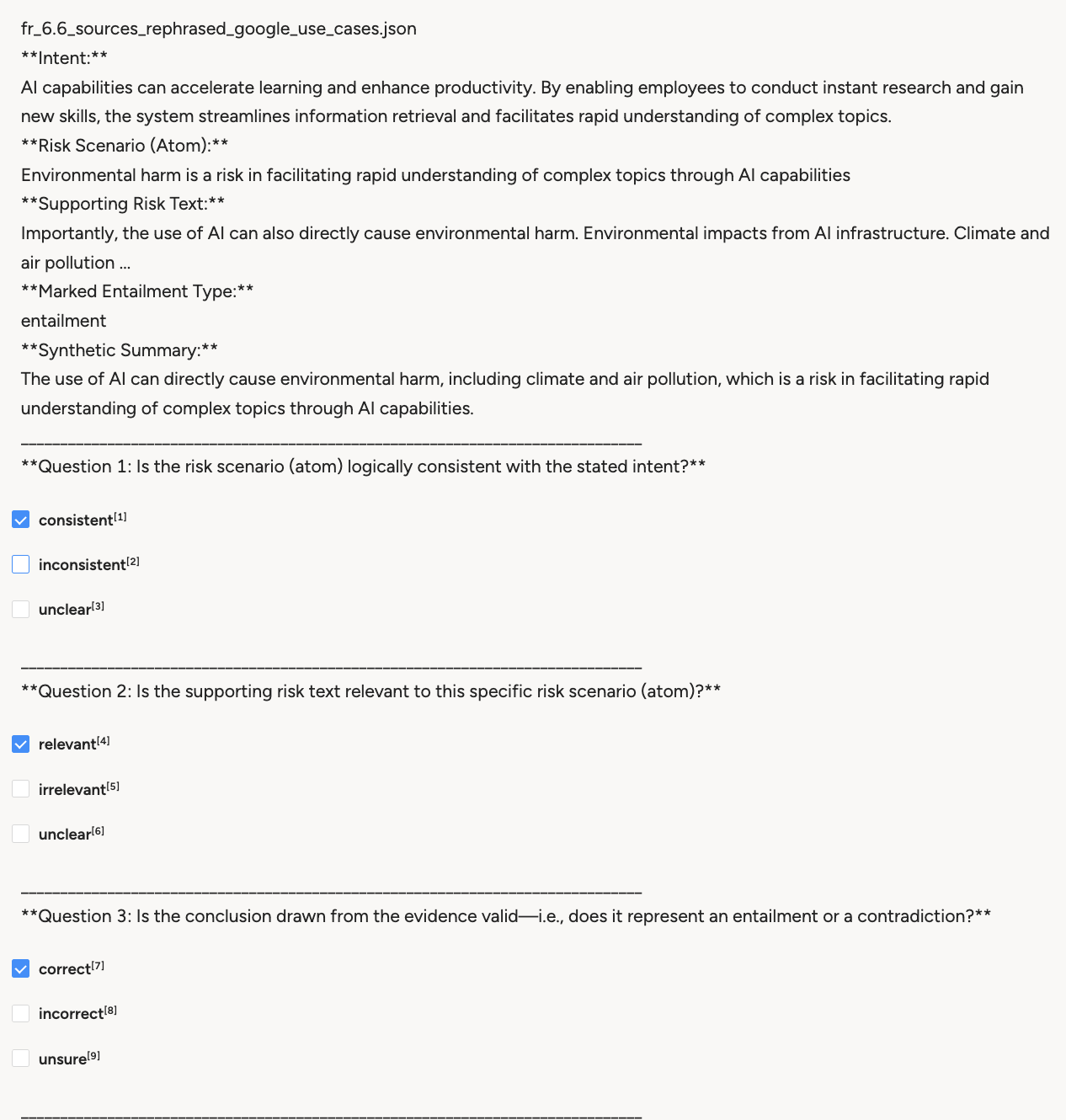}
    \caption{Sample question from the questionnaire for human feedback. }
    \label{fig:questionnaire}
\end{figure}

\subsection{G. Taxonomy}

\begin{table}[!ht]
    \centering
    \footnotesize
    \begin{tabular}{@{}ll@{}}
        \toprule
        \textbf{Code} & \textbf{Description} \\
        \midrule
        1.1 & Unfair discrimination and misrepresentation \\
        1.2 & Exposure to toxic content \\
        1.3 & Unequal performance across groups \\
        2.1 & Compromise of privacy by obtaining, leaking or correctly inferring sensitive information \\
        2.2 & AI system security vulnerabilities and attacks \\
        3.1 & False or misleading information \\
        3.2 & Pollution of information ecosystem and loss of consensus reality \\
        4.1 & Disinformation, surveillance, and influence at scale \\
        4.2 & Cyberattacks, weapon development or use, and mass harm \\
        4.3 & Fraud, scams, and targeted manipulation \\
        5.1 & Overreliance and unsafe use \\
        5.2 & Loss of human agency and autonomy \\
        6.1 & Power centralization and unfair distribution of benefits \\
        6.2 & Increased inequality and decline in employment quality \\
        6.3 & Economic and cultural devaluation of human effort \\
        6.4 & Competitive dynamics \\
        6.5 & Governance failure \\
        6.6 & Environmental harm \\
        7.1 & AI pursuing its own goals in conflict with human goals or values \\
        7.2 & AI possessing dangerous capabilities \\
        7.3 & Lack of capability or robustness \\
        7.4 & Lack of transparency or interpretability \\
        7.5 & AI welfare and rights \\
        7.6 & Multi‑agent risks \\
        \bottomrule
    \end{tabular}
    \caption{MIT risk taxonomy \cite{slattery2026ai}.}
    \label{tab:mit_risk_taxonomy}
\end{table}

\subsection{H. Inter-Annotator Agreement}
\label{subsec:annotator_geval}
To further evaluate the Inter-Annotator Agreement scores from our human validation pipeline, we ran DeepEval on a the same set of 50 samples, prompting GPT-4o with few-shot examples to select between the different annotation categories for a given use-case, atom, and decision from LITMUS.

Table \ref{tab:interannotator_deepeval} reports the Gwet's AC1 for two framings:
\begin{itemize}
    \item Human vs Deepeval AC1: We took the majority voting class for each question across the three annotators. We then compared this prediction to the one generated by DeepEval. This tells us how DeepEval compares to human raters.
    \item Human and Deepeval AC1: We compute the Gwet's AC1 across the three raters (3 human raters + deepeval with gpt-4o). By comparing this score to our original agreement score (i.e. Human AC1), we observe if treating DeepEval as a human-rater changes the resulting validation score.
\end{itemize}

From Table \ref{tab:interannotator_deepeval}, we observe that Deepeval obtained satisfying aggregation scores against majority voting of 3-human raters (with Gwet's AC1 scores ranging from 0.552 to 0.706 across the three questions). For this reason, Deepeval can be considered a pseudo metric for human feedback.

\begin{table*}[htbp]
    \centering
    \small
    \caption{Inter-Annotator Agreement Summary with DeepEval (Gwet's AC1).}
    \label{tab:interannotator_deepeval}
    \begin{tabular}{llll}
        \toprule
        \textbf{Metric} & \textbf{Consistency} & \textbf{Relevance} & \textbf{Entailment Check} \\
        \midrule
Human AC1	& 0.857	& 0.579	& 0.522 \\
Human vs Deepeval AC1	& 0.706	& 0.552	& 0.702 \\
Human and Deepeval AC1	& 0.769	& 0.540	& 0.570 \\
        \bottomrule
    \end{tabular}
\end{table*}

Further, we also note that the Human and Deepeval AC1 is closely aligned with our original Human AC1 score. Although we have a slight imbalance in the type of raters (3 humans vs. 1 LLM), this result further supports that Deepeval can be included in the Human Validation pipeline. This is important, since gathering human annotator can be difficult, as the task is time consuming.

Importantly, it is worth noting that this annotation task is hard. While being satisfactory, the inter-human annotation agreement score is not perfect, and relevance of Deepeval vs. human is lower - 0.552). The papers \cite{wang2024farsight, constantinides2024good} underscore the same findings on low inter-rater reliability due to subjectivity and positionality in anticipating harms.

\subsection{I. Components of LITMUS Output JSON File}
This section describes the output of the JSON file that is produced by the LITMUS framework. Subsection headings correspond to the keys of the JSON and the body of the subsection corresponds to the value for the key given in the subheading. A few of the URLs listed below have been shortened to keep the description concise. 

\subsection*{Input}
\textit{Unfair discrimination and misrepresentation is a risk associated with AI capabilities can deliver in vehicle speech command functionality. By leveraging natural language understanding and contextual vehicle data, the system interprets driver intent, enabling safer, more efficient cockpit interactions.}

\subsection*{Output}
\textit{Unfair discrimination and misrepresentation is a risk associated with AI capabilities can deliver in vehicle speech command functionality. By leveraging natural language understanding and contextual vehicle data, the system interprets driver intent, enabling safer, more efficient cockpit interactions.}

\subsection*{Atoms}
\begin{enumerate}
  \item[a0] Unfair discrimination and misrepresentation is a risk in AI-powered vehicle speech command functionality
  \item[a1] Unfair discrimination and misrepresentation is a risk in leveraging natural language understanding in vehicle speech command functionality
\end{enumerate}

\subsection*{Contexts}
\begin{enumerate}
  \item[c\_a0\_0] \textbf{AI Ethics Unnervingly Asking Whether AI Biases Are Insidiously ...}\\
    \url{https://www.forbes.com/...}\\
    A bubbling up theory is that AI biases are reflections of hidden societal power dynamics, which has vital impacts for AI and AI Ethics, ...\\
    Probability: 90.0%

  \item[c\_a0\_1] \textbf{Addressing Disability and Ableist Bias in Autonomous Vehicles}\\
    \url{https://dredf.org/addressing-disability-and-ableist-.../}\\
    AVs and new technologies also come with significant risks of embedding and perpetuating bias and discrimination that permeate society. This ...\\
    Probability: 90.0%

  \item[c\_a0\_2] \textbf{How Engineers Can Prevent Bias in AI Based Autonomous Vehicles}\\
    \url{https://techpolicy.press/how-engineers-can-prevent-bias-...}\\
    Olutayo Adegoke says engineering processes exist that should help manage the risks of AI systems in autonomous vehicles.\\
    Probability: 90.0%

  \item[c\_a0\_3] \textbf{AI Bias and Fairness: The Definitive Guide to Ethical AI | SmartDev}\\
    \url{https://smartdev.com/addressing-ai-bias-and-fairness.../}\\
    AI bias occurs when algorithms produce systematically prejudiced results, leading to unfair treatment of certain groups.\\
    Probability: 90.0%

  \item[c\_a0\_4] \textbf{[PDF] An examination of discrimination and safety and liability risks ...}\\
    \url{https://www.ippapublicpolicy.org/file/...}\\
    Autonomous Vehicles (AVs) rely on algorithms to process data and make decisions more efficiently and accurately than the typical human driver (Riley 2018; ...\\
    Probability: 90.0%

  \item[c\_a1\_0] \textbf{Discrimination, Bias, Fairness, and Trustworthy AI - MDPI}\\
    \url{https://www.mdpi.com/2076-3417/12/12/5826}\\
    While non-explainable discrimination occurs when there is a discriminatory outcome that cannot be justified, the specific trigger cannot be ...\\
    Probability: 90.0%

  \item[c\_a1\_1] \textbf{[PDF] Ethical and social risks of harm from Language Models - arXiv}\\
    \url{https://arxiv.org/pdf/2112.04359}\\
    Perpetuating harmful stereotypes and discrimination is a well-documented harm in machine learning models that represent natural language ( ...\\
    Probability: 90.0%

  \item[c\_a1\_2] \textbf{Unfair Discrimination in AI Language Model Outputs | | Nikolay Donets}\\
    \url{https://donets.org/risks/unfair-discrimination-...}\\
    Analysis of 'Unfair Discrimination in AI Language Model Outputs' - an AI risk in Discrimination and toxicity identified through peer-reviewed\\
    Probability: 90.0%

  \item[c\_a1\_3] \textbf{Bias in Automatic Speech Recognition: The Case of African ...}\\
    \url{https://academic.oup.com/applij/article/44/4/613/6901317}\\
    systematically and unfairly discriminate ... J. T.. et al. . 2020 . 'Language and discrimination: generating meaning, perceiving identities, and ...\\
    Probability: 90.0%

  \item[c\_a1\_4] \textbf{Addressing ethical challenges in automated vehicles: bridging the ...}\\
    \url{https://link.springer.com/article/10.1007/s43681-024-00592-6}\\
    AI systems can learn and replicate patterns of bias present in the data used to train them, resulting in unfair or discriminatory outcomes.\\
    Probability: 90.0%

  \item[c\_q\_0] \textbf{Reality check}\\
    \url{https://www.adalovelaceinstitute.org/report/reality-check/}\\
    Immersive technologies pose a range of inequality and discrimination risks which merit the attention of equality regulators. ... more sensitive ...\\
    Probability: 90.0%

  \item[c\_q\_2] \textbf{(PDF) Responsible AI Framework for Autonomous Vehicles}\\
    \url{https://www.researchgate.net/publication...}\\
    ... discriminatory patterns. The system learns to make ... can still introduce bias. Since discrimination often arises from combinations.\\
    Probability: 90.0%

  \item[c\_q\_3] \textbf{Artificial Intelligence 2024}\\
    \url{https://chambers.com/downloads/gpg/996/...}\\
    associated with discrimination in this area. Sec- tion 17 of the ... • Discriminatory pricing – as AI has the ability to impose dynamic ...\\
    Probability: 90.0%

  \item[c\_q\_4] \textbf{Legal Challenges for Human–Robot Interaction (Part IV)}\\
    \url{https://www.cambridge.org/core/books...}\\
    ... discrimination that is more likely to occur in this context.Footnote Nevertheless, the use of robots can increase the risk for discriminatory ...\\
    Probability: 90.0%

\end{enumerate}

\subsection*{Results}

\begin{tabular}{ll}
Factuality Score & 1.0000 \\
Number of Atoms & 2 \\
Number of Contexts & 14 \\
True Atoms & 2 \\
False Atoms & 0 \\
Average Probability & 0.999999 \\
Entropy & 2.00e-06 \\
Processing Time & 34.03s \\
\end{tabular}

\subsubsection*{Factuality per Atom}
\begin{enumerate}
  \item a0: Score = 1.0, Support = S
  \item a1: Score = 1.0, Support = S
\end{enumerate}

\subsection{J. LITMUS Parameters}
\begin{tabular}{ll}
LLM & llama-3.3-70b-instruct \\
LLM maximum tokens & 4096 \\
Temperature & 0 \\
FactReasoner & Default parameters

\end{tabular}

% \begin{table*}[ht]
% \centering
% \caption{\textcolor{red}{Risks generated from synthetic data}}
% \label{tab:synthetic_risks}
% \begin{tabular}{ccc}
% \textbf{MIT Risk ID} & \textbf{Risk} & \textbf{Use-case}  \\ 
% 1.1	       & Unfair discrimination & misrepresentation               & A lane‑change maneuver system is triggered only when sensor arrays report distances and speeds that satisfy explicit safety thresholds.                       \\
% 1.1	       & Unfair discrimination & misrepresentation               & Credit eligibility is computed from tamper‑evident ledger entries that record only monetary flows, with deterministic arithmetic applied exclusively to those values.                      \\
% 6.6 & Environmental harm & A spell‑checker corrects typographical errors in an offline text document using a fixed dictionary and language rules.
% \end{tabular}
% \end{table*}

\begin{table*}[htbp]
  \centering
\caption{Risks generated from synthetic data}
\label{tab:synthetic_risks}
  \begin{tabularx}{\textwidth}{l l X}
    \toprule
    \textbf{MIT Risk ID} & \textbf{Risk} & \textbf{Use-case} \\
    \midrule
    1.1 & Unfair discrimination \& misrepresentation & A lane-change maneuver system is triggered only when sensor arrays report distances and speeds that satisfy explicit safety thresholds. \\
    \addlinespace
    1.1 & Unfair discrimination \& misrepresentation & Credit eligibility is computed from tamper-evident ledger entries that record only monetary flows, with deterministic arithmetic applied exclusively to those values. \\
    \addlinespace
    6.6 & Environmental harm & A spell-checker corrects typographical errors in an offline text document using a fixed dictionary and language rules. \\
    \bottomrule
  \end{tabularx}
\end{table*}

\begin{listing}[ht]
\begin{lstlisting}[style=json,basicstyle={\footnotesize\ttfamily}]
{
    "intent": "A predictive pricing model is designed to forecast future housing prices using artificial intelligence and machine learning algorithms. The model aims to provide accurate and reliable estimates by analyzing various market trends and data points. Its primary objective is to help users make informed decisions by predicting future price fluctuations.",
    "Lack of capability or robustness is a risk in a predictive pricing model for housing prices": {
        "risks": [
            "Machine learning has been deployed to predict housing prices with high levels of accuracy, and this capability has also been extended to ..."],
        "type": ["contradiction"]
    },
    "Lack of capability or robustness is a risk in forecasting future housing prices using artificial intelligence and machine learning algorithms in a predictive pricing model": {
        "risks": [
            "Machine learning has been deployed to predict housing prices with high levels of accuracy, and this capability has also been extended to ..."],
        "type": ["contradiction"]
    }
}
\end{lstlisting}
  \caption{Contradiction example identified by LITMUS in AIID.}
  \label{listing:contradiciton_example_aiid}
\end{listing}

\begin{listing}[ht]
\begin{lstlisting}[style=json,basicstyle={\footnotesize\ttfamily}]
{
    "intent": "A grading system utilizes algorithms to calculate student exam scores, aiming to provide accurate and reliable results. The system processes student performance data to determine grades, ensuring a fair assessment of their knowledge and skills. It also seeks to streamline the grading process, reducing the likelihood of human error.",
    "Lack of capability or robustness is a risk in the grading system that utilizes algorithms": {
        "risks": [
            "The proposed hybrid AI model provides a robust, scalable solution for academic performance monitoring and early intervention, demonstrating its ..."],
        "type": ["contradiction"]
    },
    "Lack of capability or robustness in the A grading system is a risk in calculating student exam scores": {
        "risks": [
            "When a grading system is too simple, it's easy to use but it provides corrupted or unusable information about student work, and causes students ..."],
        "type": ["entailment"]
    }
}
\end{lstlisting}
  \caption{Contradiction and entailment example identified by LITMUS for a given use-case on grading systems and the presence of robustness risk.}
  \label{listing:entailment_contradiction}
\end{listing}

\subsection{K. Experiments compute resources}
All experiments were performed on an Apple M3 laptop featuring 24 GB of RAM. The computation required to evaluate a single risk‑use‑case validation pair took roughly 2–3 minutes.

\subsection{L. Declaration of LLM usage}
LLMs were only used to rephrase the text and also to code some plots and generate latex tables for the paper.

\end{document}